\documentclass[sigconf]{acmart}

\usepackage{algorithm}
\usepackage{subfigure}
\usepackage{multirow}
\usepackage{CJKutf8}
\usepackage[noend]{algpseudocode}
\AtBeginDocument{%
  \providecommand\BibTeX{{%
    \normalfont B\kern-0.5em{\scshape i\kern-0.25em b}\kern-0.8em\TeX}}}

\setcopyright{acmcopyright}
\copyrightyear{2020}
\acmYear{2020}

\acmConference[WSDM '20]{WSDM '20}{February 3--7, 2020}{Houston}
\acmBooktitle{WSDM '20:  ,
  February 3--7, 2020, Houston}

\begin{document}

\title{Conceptualized Representation Learning  for Chinese Biomedical Text Mining}


\author{Ningyu Zhang}
\email{ningyu.zny@alibaba-inc.com}
\affiliation{%
  \institution{Alibaba Group}
  \country{China}
}
\author{Qianghuai Jia}
\email{qianghuai.jqh@alibaba-inc.com}
\affiliation{%
  \institution{Alibaba Group}
  \country{China}
}
\author{Kangping Yin}
\email{kangping.yinkp@alibaba-inc.com}
\affiliation{%
  \institution{Alibaba Group}
  \country{China}
}
\author{Liang Dong}
\email{liang.dongl1@alibaba-inc.com}
\affiliation{%
  \institution{Alibaba Group}
  \country{China}
}

\author{Feng Gao}
\email{gf142364@alibaba-inc.com}
\affiliation{%
  \institution{Alibaba Group}
  \country{China}
}

\author{Nengwei Hua}
\email{nengwei.huanw@alibaba-inc.com}
\affiliation{%
  \institution{Alibaba Group}
  \country{China}
}



\begin{abstract}
Biomedical text mining is becoming increasingly important as the number of biomedical documents and web data rapidly grows. Recently, word representation models such as  BERT has gained popularity among researchers. However, it is difficult to estimate their performance on datasets containing biomedical texts as the word distributions of general and biomedical corpora are quite different. Moreover, the medical domain has long-tail concepts and terminologies that are difficult to be learned via language models. For the Chinese biomedical text, it is more difficult due to its complex structure and the variety of phrase combinations. In this paper, we investigate how the recently introduced pre-trained language model BERT can be adapted for Chinese biomedical corpora and propose a novel conceptualized representation learning approach.  We also release a new  Chinese Biomedical Language Understanding Evaluation benchmark (\textbf{ChineseBLUE}).  We examine the effectiveness of Chinese pre-trained models: BERT, BERT-wwm, RoBERTa, and our approach. Experimental results on the benchmark show that our approach could bring significant gain. We release the pre-trained model on GitHub: \url{https://github.com/alibaba-research/ChineseBLUE}. 
\end{abstract}


\maketitle

\section{Introduction}
Nowadays, the volume of biomedical literature and biomedical web pages continues to increase rapidly.  Lots of new articles and web pages containing biomedical discoveries and new insights are continuously published.  Indeed, there is an increasingly high demand for biomedical text mining.  
 \begin{figure*}
    \centering
    \includegraphics [scale=0.39]{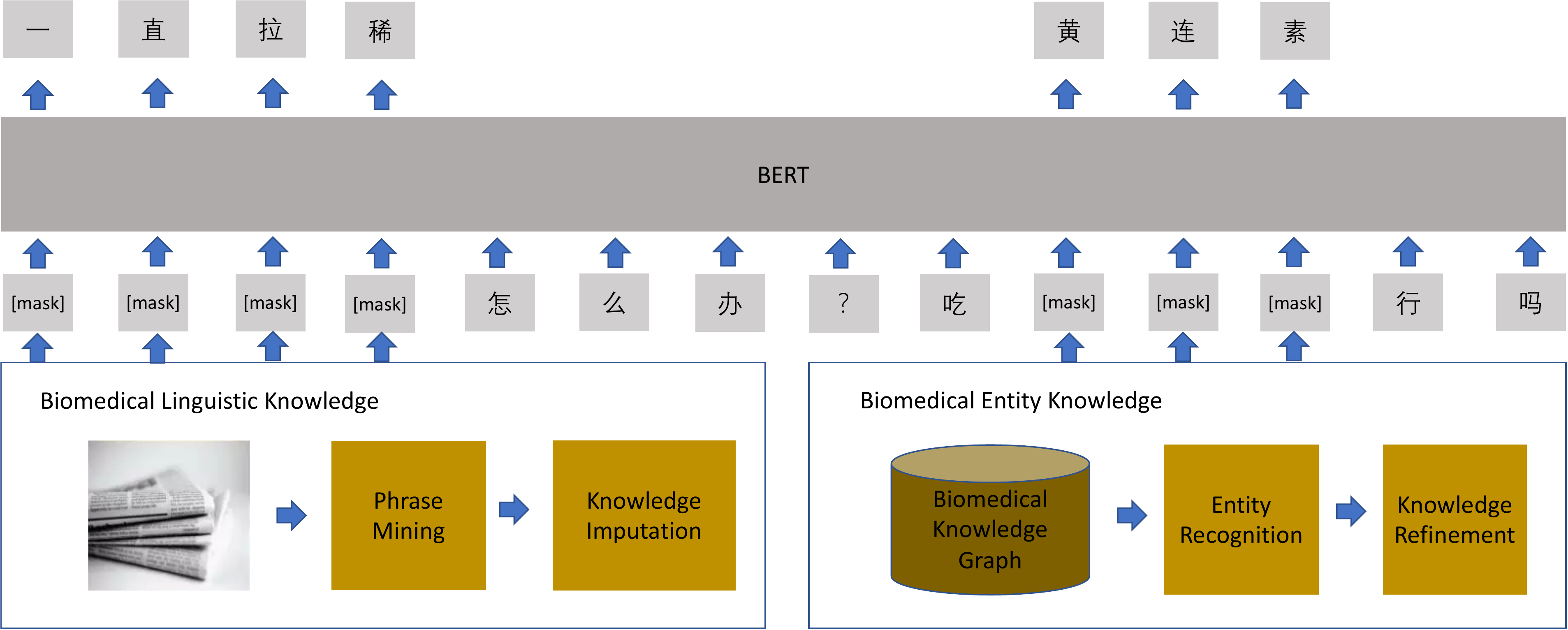}
    \caption{The structure of the conceptualized representation learning (MC-BERT) model. We utilize whole entity masking which masks the  entity "berberine" and  whole span  masking which masks the phrase "often diarrhea" to  explicitly inject biomedical knowledge.}
    \label{fig:my_label}
\end{figure*}
Recent progress in the biomedical text mining approach is made possible by the development of deep learning techniques used in natural language processing (NLP). For example, pre-trained language models such as   BERT \cite{devlin2018bert}, ERNIE \cite{sun2019ernie}, XLNet \cite{yang2019xlnet} and  RoBERTa \cite{liu2019roberta} have demonstrated remarkable successes in modeling contextualized word representations by utilizing the massive amount of training text. As a fundamental technique in natural language processing (NLP), the language models pre-trained on text could be easily transferred to learn downstream NLP tasks with finetuning, which achieve the state-of-the-art performances on many tasks including named entity recognition, paraphrase identification, question answering and information retrieval. 

However,  it has limitations to apply state-of-the-art NLP methodologies to biomedical text mining directly. Firstly, since recent representation models such as BERT  are trained and tested mainly on general domain datasets such as  Wikipedia, it is difficult to adapt to biomedical datasets without losing the performance. Moreover, the word distributions of general and biomedical text are quite different, which can be a problem for biomedical text mining. In addition, there exist long-tail concepts and terminologies in biomedical texts which are difficult to be learned via language models. For the Chinese biomedical text, it is somewhat more difficult due to its complex structure and the variety of phrase combinations. To this end, recent biomedical text mining models rely mostly on adapted versions of word representations  \cite{habibi2017deep}.  Considering whether it is possible to automatically inject biomedical knowledge to the language representation learning for Chinese medical corpus, we hypothesize that current state-of-the-art word representation models such as BERT should be trained on biomedical corpora with prior biomedical knowledge to be effective in biomedical text mining tasks. However,  there exist two problems: (1) how to retrieve the biomedical domain knowledge; (2) how to leverage such knowledge to the representation learning. 

In this paper, we propose a conceptualize representation learning approach (MC-BERT)  for Chinese biomedical language understanding. Specifically, we propose coarse-to-fine masking strategies to inject entity and linguistic domain knowledge into representation learning. As there are no benchmarks for the Chinese Biomedical Language Understanding Evaluation, we release the first large scale benchmark \footnote{\url{https://github.com/alibaba-research/ChineseBLUE}} including name entity recognition, paraphrase identification, question answering, information retrieval, intent detection, and text classification.  Experiments show that our approach achieves state-of-art results. 
\section{Related Work}
It is effective to learn general language representation by pre-training a language model with a large amount of unannotated data.  Recently, several studies BERT \cite{cui2019pre}, BERT-wwm \cite{cui2019pre},  RoBERTa \cite{liu2019roberta}, XLNet \cite{yang2019xlnet} which centered on contextualized language representations have been proposed and context-dependent language representations have shown state-of-the-art results in various natural language processing tasks. Moreover, recent studies \cite{DBLP:conf/emnlp/PetersNLSJSS19} have shown that injecting extra knowledge information can significantly enhance original models, such as knowledge acquisition \cite{zhang2019gcn}.  Some work has attempted to joint representation learning of words and entities for effectively leveraging external Knowledge Graphs and achieved promising results \cite{zhang2019improving,taolin2020}. \cite{sun2019ernie} propose the knowledge masking strategy for masked language models to enhance language representation by knowledge.  However, there is only a little research on pretraining in the biomedical domain. \cite{lee2019biobert} firstly proposed   BioBERT, which is a domain-specific language representation model pre-trained on large-scale biomedical corpora. In this paper, we further leverage both corpora and knowledge graphs to train an enhanced language representation model based on BERT. To the best of our knowledge, we are the first approach to inject biomedical knowledge into Chinese biomedical representation leaning. 

\section{Approach}
We present MC-BERT, a self-supervised pretraining method designed to represent the text better. Our approach is inspired by BERT but deviates from its bi-text classification framework in three ways. First, we use a different mask generation procedure to mask spans of tokens, rather than only random ones. We also introduce two kinds of masking strategies, namely whole entity masking and whole span masking.  Finally, MC-BERT split the input document into segments based on the actual "sentences" provided by the user as positive samples and sample random sentences from other documents as negative samples for the next sentence prediction.  
\subsection{Whole Entity   Masking}
Different from whole word masking, we do not mask random words but medical entities such as "\begin{CJK}{UTF8}{gbsn}腹痛\end{CJK}" ("stomach ache"), which can explicitly inject medical domain knowledge.  We utilize the Chinese biomedical knowledge graph and biomedical named entity recognition to extract and refine entities in the medical domains, including syndrome, decease,  examination, treatment, drug, and so on. 

\subsection{Whole Span  Masking}
 Due to the complex structure and the variety of phrase combinations in Chinese, we also inject fine-grained biomedical knowledge with whole span masking.   For example, the phrases "\begin{CJK}{UTF8}{gbsn}肚子有一点疼\end{CJK}" ("a little pain in the stomach"), “\begin{CJK}{UTF8}{gbsn}腹部一阵一阵痛\end{CJK}” ("a pain in the abdomen"), “\begin{CJK}{UTF8}{gbsn}腹痛\end{CJK}” ("stomach ache") have the same meaning with concepts "\begin{CJK}{UTF8}{gbsn}腹痛\end{CJK}" ("stomach ache"), the whole entity masking cannot explicitly inject such knowledge.  We argue that the prior linguistic knowledge of phrases is necessary for Chinese biomedical language understanding. Specifically, we firstly utilize Autophrase\footnote{\url{https://github.com/shangjingbo1226/AutoPhrase}} to extract phrases. We also retrieve common biomedical phrases  from Alibaba Congitive Concept Graph\footnote{\url{https://github.com/alibaba-research/CognitiveConceptGraph}}.  Then we leverage domain rules to augment data and train a binary classifier to filter those none-biomedical phrases.  We collect the n-gram of entities and attribute words in the medical encyclopedia as positive samples and generate random phrases as negative samples for the classifier via fastText\footnote{\url{https://github.com/facebookresearch/fastText}}. 
  
Note that all of the words in the same unit are masked during word representation training instead of only one word or character being masked. In this way, the prior knowledge of phrases and entities are explicitly learned during the training procedure. Instead of adding the knowledge embedding directly, MC-BERT explicitly learned the information about knowledge to guide word embedding learning. This can make the model have better generalization and adaptability in the biomedical domain. 

\subsection{Further Pretraining in Biomedical Domain}
We did not train our model from scratch but from the BERT-base. We trained 100K steps on the samples with a maximum length of 512, with an initial learning rate of 1e-5\footnote{The initial learning rate should be adapted for different corpus}. Note that the learning rate is critical, and we \textbf{donnot} use the learning rate warmup as we empirically find it will lead to severe catastrophy forgetting in the biomedical domain.  The overall training procedure is shown below.


\begin{algorithm}[th]
\caption{Overall Training Procedure of MC-BERT}
\label{algorithm: training}
\begin{algorithmic}[1]
\State Generate candidate biomedical entities and refine them from the biomedical knowledge graph.  
\State  Generate candidate phrases via Autophrase from the raw corpus.

\State  Augment and filter those phrases via rules and fastText.

\State  Duplicate and shuffle corpus ten times and generate whole entity/span masking training samples with rate 15\%.

\State Initialize all parameters with BERT-base and make further pretraining in the biomedical corpus.

\end{algorithmic}
\end{algorithm}

\section{Experiments}
\subsection{Pretraining Data and Settings}
For the Chinese corpus, we collect a variety of data, such as Chinese community biomedical question answering, Chinese medical encyclopedia, Chinese electronic health records (EHR), and so on from Alibaba Shenma Search Engine\footnote{\url{http://m.sm.cn}}. The details of the pretraining corpus are shown in Table \ref{pretraining}.

\begin{table}
\caption{Statistics of pretraining corpus.}
\begin{tabular}{c|c}
\hline
\hline
Corpus Type&\#Sentences \\
\hline
Chinese Biomedical Community QA&20M \\
Chinese Medical Encyclopedia&100K \\
Chinese Electric Medical Record&10K \\
\hline
\hline
\end{tabular}\label{pretraining}
\end{table}

\begin{table}
\caption{Statistics of ChineseGLUE.}
\begin{tabular}{c|ccccc}
\hline
\hline
Dataset&Train&Dev&Test&Task&Metric \\
\hline
cEHRNER&914&44&41&NER&F1\\        
cMedQANER&1,673&175&215&NER&F1\\        
cMedQQ&16,071&1,793&1,935&PI&F1 \\   
cMedQNLI&80,950&9,065&9,969&QA&F1\\  
cMedQA&49,719&5,475&6,149&QA&F1 \\
cMedIR&80,000&10,000&10,000&IR&PAIR\\
cMedIC&1,683&123&84&IC&F1 \\
cMedTC&14,610&1,550&1,800&TC&F1 \\
\hline
\hline
\end{tabular}\label{chineseglue}
\end{table}
 
To compare with BERT, we leverage the same model settings of the transformer as BERT. The base model contains 12 layers, 12 self-attention heads, and 768-dimensional of hidden size, while the large model contains 24 layers, 16 self-attention heads, and 1024-dimensional of hidden size. 

\subsection{Finetuning Tasks}
We executed extensive experiments on  Chinese NLP tasks and release a Chinese Biomedical Language Understanding Evaluation benchmark (ChineseBLUE), as shown in table \ref{chineseglue}. The following Chinese datasets in the ChineseBLUE are chosen to evaluate the performance of MC-BERT on Chinese tasks: 
\begin{itemize}
\item \textbf{Named Entity Recognition (NER)} aims to recognize various entities, including diseases, drugs, syndromes, etc.   The cEHRNER dataset labeled from the Chinese electronic health records and the cMedQANER dataset labeled from Chinese community question answering is chosen.

\item \textbf{Paraphrase Identification (PI)} aims to identify whether two sentences express the same meaning. We use cMedQQ, which consists of search query pairs. 
 
\item \textbf{Question Answering (QA)}, which can be approximated as ranking candidate answer sentences based on their similarity. We assign 0,1 labels to the QA pairs, which convert to the binary classification problem. We use cMedQNLI, which consists of long answers and cMeQA, which consists of short answers. 

\item \textbf{Information  Retrieval (IR)}, which aims to retrieve most related documents given search queries. IR can be regarded as a ranking task. We adopt the \textbf{PAIR}\footnote{A popular NDCG-like ranking metric in the search engine, which refers to the number of positive ranked documents divide the number of negative ranked documents.} score to evaluate the model.  We use the cMedIR dataset,  which consists of queries with multiple documents and their relative scores. 

\item \textbf{Intent Classification (IC)} aims to assign intent labels to the queries, which can be regarded as multiple label classification tasks. We use the cMedIC dataset, which consists of queries with three intent labels (e.g., no intention, weak intention, and firm intention).

\item \textbf{Text Classification (TC)} aims to assign multiple labels to the sentence. We use the cMedTC dataset, which consists of biomedical texts with multiple labels.

\end{itemize}

\begin{table}
\caption{BERT and MC-BERT results on different tasks of ChineseBLUE.}
\begin{tabular}{c|cccc}
\hline
\hline
Model&cEHRNER&cMedQANER&cMedQQ&cMedQA \\
\hline
BERT-base&88.2&86.3 &86.5&81.0\\
MC-BERT&\textbf{90.0}&\textbf{88.1}  & \textbf{87.5} &\textbf{82.3}\\
\hline
\hline
Model&cMedQNLI&cMedIR&cMedIC&cMedTC\\
\hline
BERT-base& 93.3 &1.77 & 86.0 &79.0  \\
MC-BERT& \textbf{95.5} & \textbf{2.04} & \textbf{87.5}  &\textbf{82.1}\\
\hline
\hline
\end{tabular}\label{cbertres}
\end{table}

\subsection{Results}

Table \ref{cbertres} shows the performances of classical Chinese NLP tasks. It can be seen that MC-BERT outperforms BERT-base on all tasks, including cEHRNER, cMedQANER, cMedQQ, cMedQA, yet the performance achieve only a little progress on the rest, which is caused by the difference in pretraining between the two methods. Specifically, the pretraining data of MC-BERT does not contain instances whose length less than  128, but the length of finetuning instances in datasets such as cMedTC is less than 128.   Specifically, MC-BERT yields improvements of more than 2 points over BERT-base on the name entity recognition, and yields improvements of more than 1 points over BERT-base on the paraphrase identification task. 

\subsection{Analysis of Chinese Pretrained Models}
To better demonstrate the performance of different strategies in our model and different pretraining approaches in Chinese, we separately remove the whole entity and span masking and also compare our model with other Chinese pretraining models.  The experimental results of named entity recognition on the cEHRNER dataset are summarized in Table~\ref{tab:ablation}. 
\textbf{MC-BERT} is our method;  
\textbf{w/o entity} is the method without whole entity masking; 
\textbf{w/o span} is the method without whole span masking; 
\textbf{BERT-wwm} \cite{cui2019pre} is a whole word masking pretraining approach for the Chinese text; \textbf{RoBERTa} \footnote{We adopt all the pretrained Chinese model from \url{https://github.com/ymcui/Chinese-BERT-wwm}} \cite{liu2019roberta} is a robustly optimized BERT pretraining approach.  We observe that: (1) Performance degrades when we remove "whole entity mask"  and "whole span mask." This is reasonable because the entity and span explicitly inject domain knowledge into representation learning.  (2) Our approach performs better than the recent state-of-art pretraining approach RoBERTa in Chinese, which indicates further pretraining in the biomedical domain is necessary.  

\begin{table}
\caption{NER Results of different  Chinese pretraining models and ablation study.}
\begin{tabular}{c|ccc}
\hline
\hline
Model&Precision&Rrecall&F1\ \\
\hline
MC-BERT&\textbf{89.8}&\textbf{90.4}&\textbf{90.0} \\
w/o entity&89.5&89.6&89.6 \\
w/o span&88.1&88.3&88.2 \\
\hline
BERT-base&88.0&88.4&88.2 \\
BERT-wwm&89.1&89.2&89.2 \\
RoBERTa&89.1&89.5&89.3 \\

\hline
\hline
\end{tabular}
\label{tab:ablation}
\end{table}
\subsection{Discussion}
As far as we concern, there are only a few Chinese open biomedical datasets that hinder the research of Chinese biomedical language understanding. We make the first step to build a Chinese Biomedical language understanding benchmark and investigate the pretraining strategies. In our experiments, we notice that there exit long-tail terminologies in Chinese biomedical corpus such as "\begin{CJK}{UTF8}{gbsn}肌酶\end{CJK}" ("muscle enzyme"). The vanilla mask language model can not learn robust representations from only word-level masking. Therefore, we propose the whole entity/span masking to inject domain knowledge explicitly, which improves the performance.  Note that our approach is language-agnostic, it is convenient to adapt our approach to other languages such as English and even low-resource languages.  However, our work has some limitations. First, our work relies on phrase mining, which makes the whole training procedure not end-to-end and may result in error propagation. Otherwise, there is also some concept knowledge in the biomedical domain, such as various phrases refer to the same normalized concept. We believe the performance of our model can be further improved by injecting such knowledge, which will be part of our future work.

\section{Conclusion and Future Work}
In this article, we introduce MC-BERT, which is a pre-trained language representation model with domain knowledge for biomedical text mining. We show that pretraining BERT on biomedical corpora is crucial in applying it to the biomedical domain. The ChineseBLUE and MC-BERT will soon be available to the BioNLP community. Our motivation is to develop a universal, GLUE-like, and open platform for the Chinese BioNLP community and a  composable and generalized representation algorithm to inject domain knowledge.  Our work is but a small step in this direction.  

\bibliography{WSDM2020}
 
\bibliographystyle{ACM-Reference-Format}


\end{document}